\documentclass[11pt,a4paper]{article}
\usepackage[hyperref]{acl2018}
\usepackage{times}
\usepackage{latexsym}
\usepackage{lipsum}
\usepackage{arydshln}
\usepackage{graphicx}
\usepackage{url}

\aclfinalcopy 

\title{Explain to me like I am five - Sentence Simplification Using Transformers}

\author{Aman Agarwal \\
  Luddy School of Informatics, Computing, and Engineering \\
  Indiana University, Bloomington, IN 47408 \\
  {\tt amanagar@iu.edu} \\}

\date{}

\begin{document}
\maketitle
\begin{abstract}
    Sentence simplification aims at making the structure of text easier to read and understand while maintaining its original meaning. This can be helpful for people with disabilities, new language learners, or those with low literacy. Simplification often involves removing difficult words and rephrasing the sentence. Previous research have focused on tackling this task by either using external linguistic databases for simplification or by using control tokens for desired fine-tuning of sentences. However, in this paper we purely use pre-trained transformer models. We experiment with a combination of GPT-2 and BERT models, achieving the best SARI score of 46.80 on the Mechanical Turk dataset, which is significantly better than previous state-of-the-art results. The code can be found at {\small \url{https://github.com/amanbasu/sentence-simplification}}.
\end{abstract}

\section{Introduction}
Sentence Simplification (SS) involves generating a simpler version of a complex sentence while maintaining its meaning. This can be helpful for people with disabilities like dyslexia, aphasia or autism, language learners, or those with low literacy. For example, aphasic patients face challenges in following long sentences with convoluted structure, or, new language learners are generally unaware of rare words and phrases. Simplification, thereby, typically involves removing unnecessary or complex words and paraphrasing for simplicity. This improves the text's accessibility and readability, making it easier for a wider audience to understand and retain the content.

In this paper, we propose an approach for the task of SS using fine-tuned Transformers~\cite{vaswani2017attention}. With the advent of Transformer models, the whole field of Natural Language Processing has been revolutionized. Unlike auto-regressive models like RNN and LSTM, these feed-forward and fully-connected models are based on the idea of self-attention, where the model attends to different parts of the input sequence to generate an output. Transformer models can be pre-trained on large corpora, such as Wikipedia or the entire Internet's text, to capture a wide range of knowledge and language patterns. Subsequently, the models can then be fine-tuned on a specific tasks, such as language translation or sentence simplification, to further improve their performance.

We use two popular transformer models, BERT \cite{devlin2018bert} and GPT-2 \cite{radford2019language}. BERT (Bidirectional Encoder Representations from Transformers) is a natural language processing (NLP) model developed by Google. It is trained in a bidirectional fashion, meaning that it can take into account the context of the words in a sentence from both the left and the right sides, rather than just the left side of a word. GPT-2 is another popular transformer model from OpenAI. The key difference between the two is that GPT-2 is a unidirectional model that only considers the left context when processing input data. This may intuitively make BERT better at understanding the meaning of words in a sentence, however, GPT-2 is faster and more efficient at generating text. 

After training these models on a huge text corpus, they could be fine-tuned to do any particular activity, such as question answering, summarizing, translating, reading comprehension, etc. We show that fine-tuning a large language model (LM) like BERT or GPT-2 is good enough for SS, rather than involving additional linguistic knowledge like used in the past. Our SARI score \cite{xu2016optimizing} of \textbf{46.80} on the Mechanical Turk dataset beats the state-of-the-art (SOTA) by a huge margin of \textbf{3.49}.

\section{Related Work}
Recent work on sentence simplification using transformer models has focused on using these models to automatically generate simpler versions of complex sentences. These models have been shown to be effective at generating simpler versions of sentences while maintaining their meaning~\cite{fang2019sentence, zhao2018integrating, martin2019controllable, omelianchuk2021text, sheang2021controllable, tajner2022SentenceSC}. 

Some of these approaches have also explored using additional linguistic knowledge to improve the quality of the generated simplified sentences. Especially, through the use of Simple Paraphrase Database (SimplePPDB) \cite{pavlick-callison-burch-2016-simple}. It contains a collection of around 4.5 million rules for text reduction in English. SimplePPDB is generally used to provide the model with a set of potential reduction rules for a given complex sentence. These rules can be incorporated in the model's loss function during training~\cite{fang2019sentence, zhao2018integrating}.

Additionally, researchers have also explored controllable sentence simplification~\cite{martin2019controllable, sheang2021controllable, tajner2022SentenceSC}. Controllable sentence simplification is the process of using control tokens while training transformer models to generate simpler versions of sentences. This allows the model to be more flexible in generating a wide range of simplifications, depending on the specific control tokens provided. For example, a model trained with control tokens for ``simplicity" and ``fluency" can generate simpler sentences with a higher level of fluency. Similarly, a model trained with control tokens for ``words" would generate sentences that use smaller, and thereby, less complicated words.

During training, the model is provided with a set of control tokens and their corresponding target simplifications. The model then learns to generate the target simplification based on the provided control tokens. During inference, the model can be provided with a new set of control tokens and the original complex sentence. The model then generates a simplification based on the control tokens and the underlying structure and meaning of the complex sentence. Overall, controllable sentence simplification using control tokens allows the model to generate a wider range of simplifications, tailored to specific control tokens and requirements.

Furthermore, some studies like \citet{omelianchuk2021text} have tried to pose simplification as a tagging problem, similar to Named Entity Recognition. The model takes a complex sentence and suggests some edits (through tags) to make it simpler. For example, convert verb to its simple/base form from its third person singular present tense form; append a word; or replace a complex word with its simpler alternative. This process is then repeated for a fixed number of times as some of the edits can be dependent on the other edits. The process eventually leads to a simpler sentence.

Overall, the use of transformer models in sentence simplification has shown promising results, with a potential for further improvement through fine-tuning, control tokens, and incorporating existing linguistic knowledge.

\section{Methods}

\subsection{Data Source}
We used the \textit{WikiLarge} dataset created by \citet{zhang2017sentence}. It was formed by merging data from three different sources, the major one being from \citet{coster2011simple}. \citet{coster2011simple} data contains 137K aligned sentence pairs from \footnote{http://en.wikipedia.org/}{English Wikipedia} and corresponding simplified sentences from \footnote{http://simple.wikipedia.org}{Simple English Wikipedia}. The total size of \textit{WikiLarge} by merging the three sources is 296K.

For the validation and test purpose, data from \textit{WikiSmall} and the simplifications provided by Amazon Mechanical Turk workers were used \cite{xu2016optimizing}. The validation and test set contains 2000 and 359 sentence pairs respectively, along with 8 reference pairs. This has sort of become a standard dataset for SS task and is used in numerous studies. Furthermore, all the studies compared in this paper use the same data.

\subsection{Model}
We have used the \textit{EncoderDecoderModel} provided by \footnote{https://huggingface.co}{HuggingFace's} transformers library with a combination of \textit{BertConfig} and \textit{GPT2Config} as encoder or decoder. The model is implemented in PyTorch and trained on Nvidia V100 GPU cluster system. BERT uses a vocabulary size of $ 30522 $ while GPT-2 uses $ 50257 $. Both of them uses embedding dimension of $ 768 $ and $ 12 $ attention heads. They have a a total of $ 12 $ hidden layers connected via GeLU non-linearity. The maximum length of the sentence tokens have been fixed at 80.

For training, AdamW optimizer was used with the initial learning rate of $ 1e-4 $. It was controlled by a one-cycle learning rate scheduler with the maximum rate of $ 1e-3 $ and warm start. We used early stopping based on the SARI score to prevent overfitting of the model. The best model required a total of 20 epochs and around 24 hours for fine-tuning. 

\subsection{Evaluation Metric}
SARI - {\bf S}ystem output {\bf a}gainst {\bf r}eferences and against the {\bf i}nput sentence, is the most common metric for evaluation of SS. It compares the simplified sentence with both source and reference sentences. Using ngrams, SARI calculates the scores for three conditions: additions, keeps, and deletions. In general, it gives a high score when a simple word is added that is absent in the source sentence but present in the reference sentence, when a complex word is removed that is present in the source sentence but absent from the reference, or when the word is retained in both the source and reference sentences. These separate scores are averaged to determine the result.

\section{Results \& Discussion}
We evaluate the results of our models on SARI and compare them against some of the past studies. SARI is an average of three components: addition, deletion and keep score as compared between the source, target and the reference sentences. We report all of these scores, see Table \ref{table:results} for details. In addition to  using BERT and GPT-2 as encoder and decoder, we also tried a mix of both. BERT+GPT-2 means that the encoder was made of BERT and the decoder was GPT-2, and vice versa for GPT-2+BERT.

\begin{figure}
	\centering
	\includegraphics[width=\linewidth]{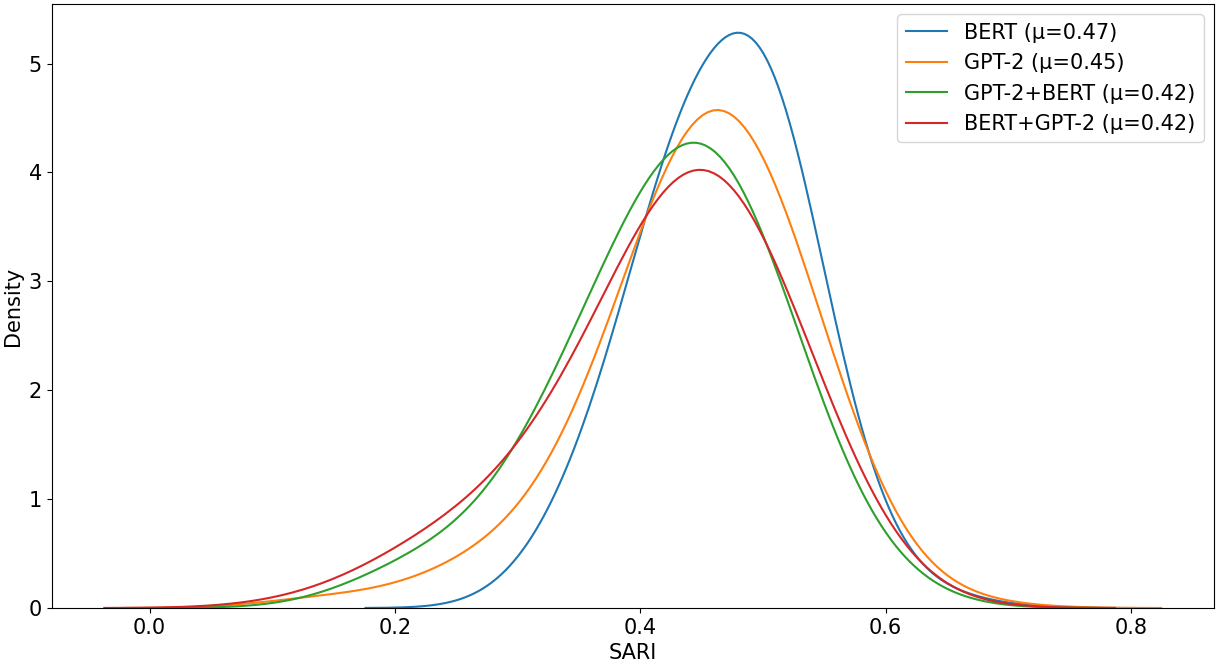}
	\caption{Distribution of SARI score for all test samples across different transformer models.}
	\label{fig:sari}
\end{figure}

With a SARI score of 46.80, BERT emerged as the top model overall. It may be as a result of the network's bidirectional nature. In order to decide whether or not to modify a word, it may be helpful to consider its context from both sides. The distribution of test scores in Figure \ref{fig:sari} also helps in showing the large number of samples getting a high SARI score of around 0.5 in BERT. Following closely after in second place was the GPT-2 model. The lower performance of GPT-2 can perhaps be due to its unidirectional nature. Lastly, the combination of BERT and GPT-2 encoder and decoder didn't seem to perform very well. It may be inferred that rather than mixing, both of these models perform best when used within the common architecture.

\begin{table}
\centering
\resizebox{\linewidth}{!}{
\begin{tabular}{l|c|ccc}
    Model & SARI $\uparrow$ & ADD $\uparrow$ & DELETE $\uparrow$ & KEEP $\uparrow$ \\
    \hline
    \citet{zhao2018integrating} & 40.42 & 5.72 & 42.23 & {\bf 73.41}  \\
    \citet{martin2019controllable} & 41.87 & - & - & - \\
    \citet{omelianchuk2021text} & 41.46 & 6.96 & 47.87 & 69.56 \\
    \citet{sheang2021controllable} & 43.31 & - & - & - \\
    \citet{tajner2022SentenceSC} & 43.30 & - & - & - \\ 
    \hdashline[3pt/2pt]
    BERT+GPT-2 & 42.31 & 11.07 & 62.82 & 53.93 \\
    GPT-2+BERT & 42.35 & 10.74 & 62.37 & 54.05 \\
    GPT-2 & 46.35 & {\bf 12.60} & 66.64 & 59.73 \\
    BERT & {\bf 46.80} & 12.13 & {\bf 67.16} & 61.22 \\
    \end{tabular}}
    \caption{Comparison of evaluation scores on Mechanical Turk dataset.}
    \label{table:results}
\end{table}

\section{Conclusion \& Future Work}

Sentence simplification is the process of creating a simpler version of a complex sentence and can be useful for people with disabilities, language learners, or those with low literacy. Simplification often involves removing complex words and paraphrasing to make the text easier to understand and read. In this paper, we propose using fine-tuned transformer models for sentence simplification. We use a combination of transformer encoder and decoder models, BERT and GPT-2 to be precise. The BERT model proved to be the best of all the models used in this study and other previous studies. A SARI score of 46.80 on the Mechanical Turk dataset beats the state-of-the-art by a huge margin. In future, we would like to explore these models on the task of \textit{controlled} sentence simplification.

\section*{Acknowledgments}
Professor Francis M. Tyers for giving this opportunity and Indiana University for providing the GPU computing resources.

\bibliography{acl2018}
\bibliographystyle{acl_natbib}

\end{document}